# Behaviora - A Conceptual Architecture for External and Internal Behavior of Robots and Agents

Göte Nyman

**Abstract**

Behaviora is a preliminary conceptual architecture for representing agent and robot behavior, external and internal alike, in an addressable form. A behaving robot or agent performs a *Behavior Episode* composed of episode components, which can be derived from behavior taxonomies (BTax) and assigned persistent identifiers. We denote these identifiers as *IoB (Internet of Behaviors) Addresses*. A *Behavior Episode* specifies what the system does, while a *Style Profile* (SP) specifies how this behavior is expressed. Style can communicate characteristics of the actor and qualities such as competence and cultural manners. An *Experience Profile* (EP) represents behaviorally relevant internal state that modulates the execution of an Episode. Finally, a *Behavior Compiler* maps these behavioral representations to platform-specific actions. We use a primitive 'touching arm' model to show these components and their relations. *External Behavior* is a result of addressable movements and their styles. *Internal Behavior* is represented through the same episodic principle and can be rendered as inner speech. Sensing, perception and complex task contexts have not been included in the present implementation, although a conceptual place is reserved for them.

## 1. Introduction

Robots and artificial agents have become increasingly diverse, and operate in complex physical, virtual, and social environments. Embodied AI systems integrate heterogeneous models and data to act intelligently across different tasks and situations. Scaling of this development is an acute challenge.

Standardization in humanoid robotics is discussed in Liu et al. [1], for example, dealing with the development of ISO/WD 26264-1 for humanoid robot datasets. They expect the future scalability of humanoid robotics to depend on making embodied episodes shareable and reusable across robots, tasks, organizations, and time. Accordingly, humanoid data should not be treated as isolated samples because it must preserve the relationships among e.g., the robot body, task, physical scene, and execution. The authors emphasize episode schemas, and specific grammars for relevant domains such as manipulation, locomotion, human–robot interaction, and decision making [1]. We share this general motivation and approach the standardization problem from a purely behavioral direction.

The common denominator across robots, agents, and their embodiments in different environments is ultimately *the behavior*. It determines how they interact with people, respond to their environments, pursue goals, and how they are understood.

A robot body may change, its sensors and actuators differ, and its underlying control architecture may be replaced, but a recognizable behavior, e.g., approaching an object, greeting a certain person, or drawing a line with a style, can remain conceptually identifiable as behaviors. Behavior is something we humans can observe and understand and it carries cultural signals and meanings.

Hence, we asked a simple question: can behavior, external and internal alike, itself become an explicitly represented and shareable computational entity? Behaviora explores this possibility as an embodiment-independent layer between intelligent processes and their platform-specific execution.

The behaviors of existing robotics and AI systems are guided by plans, rules, movement trajectories and sequences, behavior trees, and learned latent structures. Amazing performances have become possible. *Behaviora* is a conceptual and preliminary framework for representing, sharing, and addressing internal and external behaviors (see Figure 1).

We make a peculiar distinction between *External Behavior* and *Internal Behavior*. External Behavior refers to behavior observable through action, movement, or any physical interaction with the environment. Internal Behavior refers to identifiable behavior episodes occurring within the behaving system, such as intention formation, memory retrieval, expectation, experiencing, attention, decision making, and self-monitoring.

We do not assume that representing such processes establishes subjective experience or consciousness similar to the human mind. Our proposal is purely computational. An identifiable internal process (episode) can be treated as behavior, and represented using the same Internet of Behaviors (IoB) [2,3] formalism as externally observable behavior. As a result, representation of Internal Behavior can be shared, edited and otherwise used for various purposes.

In Behaviora every behavior is represented as a coherent Behavior Episode, which consists of a sequence of episode components or sub-episodes. These and the main episode itself have persistent hierarchical IoB Addresses. A Behavior Episode represents what a robot or agent does.

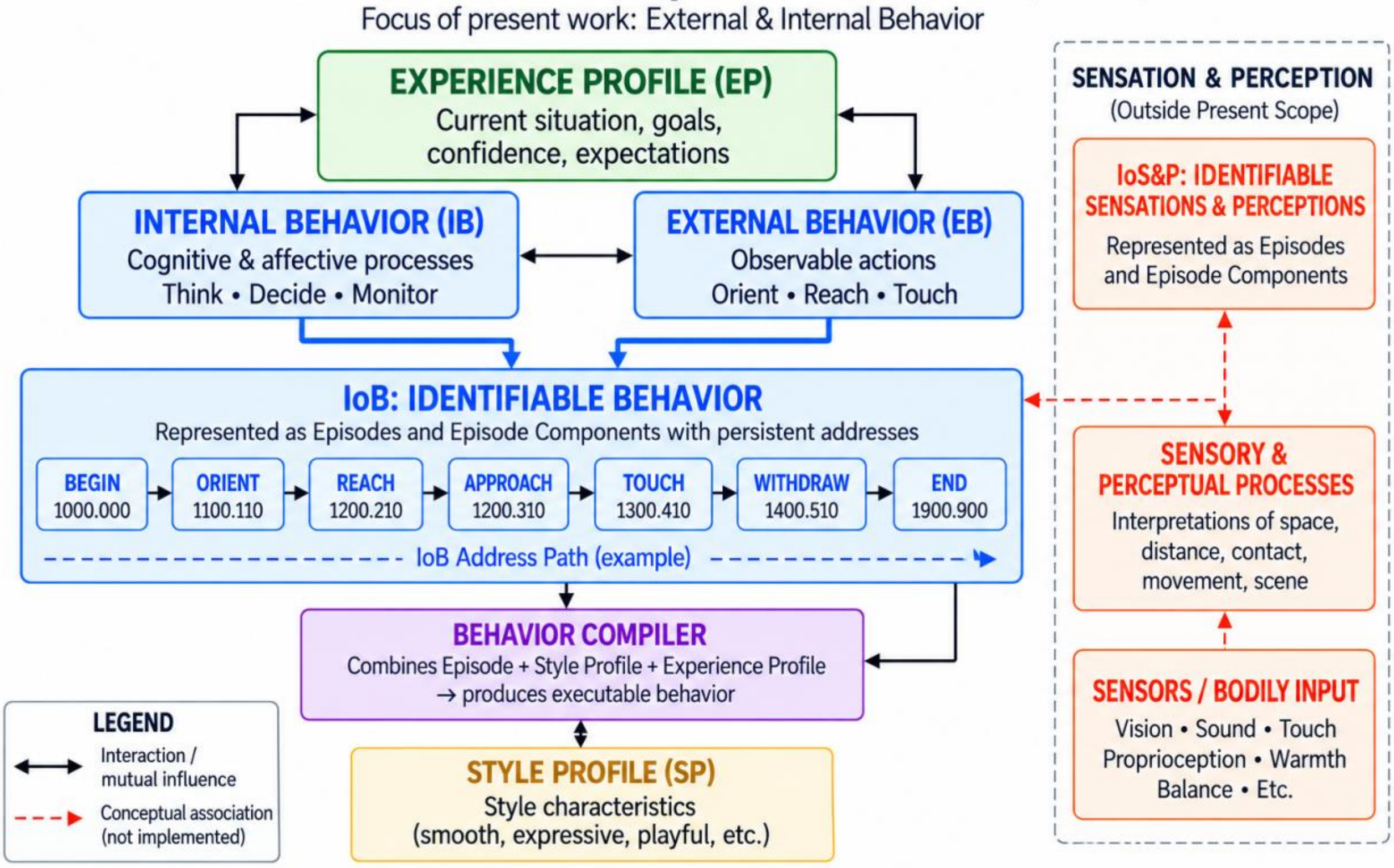

**Figure 1. Behaviora: Conceptual architecture**

*A sequence of IoB addresses identifies the components of a Behavior Episode. Style Profile and Experience Profile act as separate modulation channels, and the Behavior Compiler realizes the resulting specification on a particular platform. Sensation and perception are shown separately as a reserved conceptual stream outside the present implementation. Task contexts have not been included in this outline.*

A Style Profile (SP) specifies *how* this behavior is expressed, while an Experience Profile (EP) represents any relevant internal state that modulates its execution. Behaviora makes such internal phenomena meaningful, explicit, and shareable objects.

A *Behavior Compiler* translates the relevant behavioral representation, Behavior Episode, together with applicable SP and EP information, into platform-specific actions required by a particular robot, software agent, or any other behaving system. As a result, behavioral identity can be preserved while its execution differs across embodiments. A physical robot and a virtual character need not possess the same morphology, control system, or actuators in order to share a (representation of a) behavior. The Compiler translates the represented Behavior Episode into an execution appropriate to its own mechanical capabilities.

We are aware of several open questions, including how to construct behavioral taxonomies, temporality, how to determine Behavior Episode boundaries, and how to map equivalent behaviors across radically different embodiments. Some of these are briefly addressed in the Discussion.

How to classify behaviors? Behavior taxonomy [4] is one way to organize behaviors in a meaningful way and have human-readable labels for them and their components. However, the verbal labels of behaviors in taxonomies do not make behaviors unique and addressable. A word can have several meanings, and different words can refer to the same behavior. Behaviora uses IoB Addresses as unique identifiers for behavioral entities. Other schemes are, of course, possible.

Established approaches address different aspects of this problem. Planning and goal-based systems organize intended action and seek to make the reasoning behind goal-directed behavior understandable [5]. XAI addresses the broader problem of making AI decisions and actions explainable to human users [6]. Behavior Trees provide modular hierarchical structures for organizing actions [7], while Belief-Desire-Intention (BDI) systems represent beliefs, desires, intentions, and plans [8]. Cognitive architectures provide broader frameworks integrating capabilities such as perception, attention, action selection, memory, learning, and reasoning [9]. Reinforcement learning acquires behavior from data, and recent vision-language-action models have also represented robot actions as tokens [10].

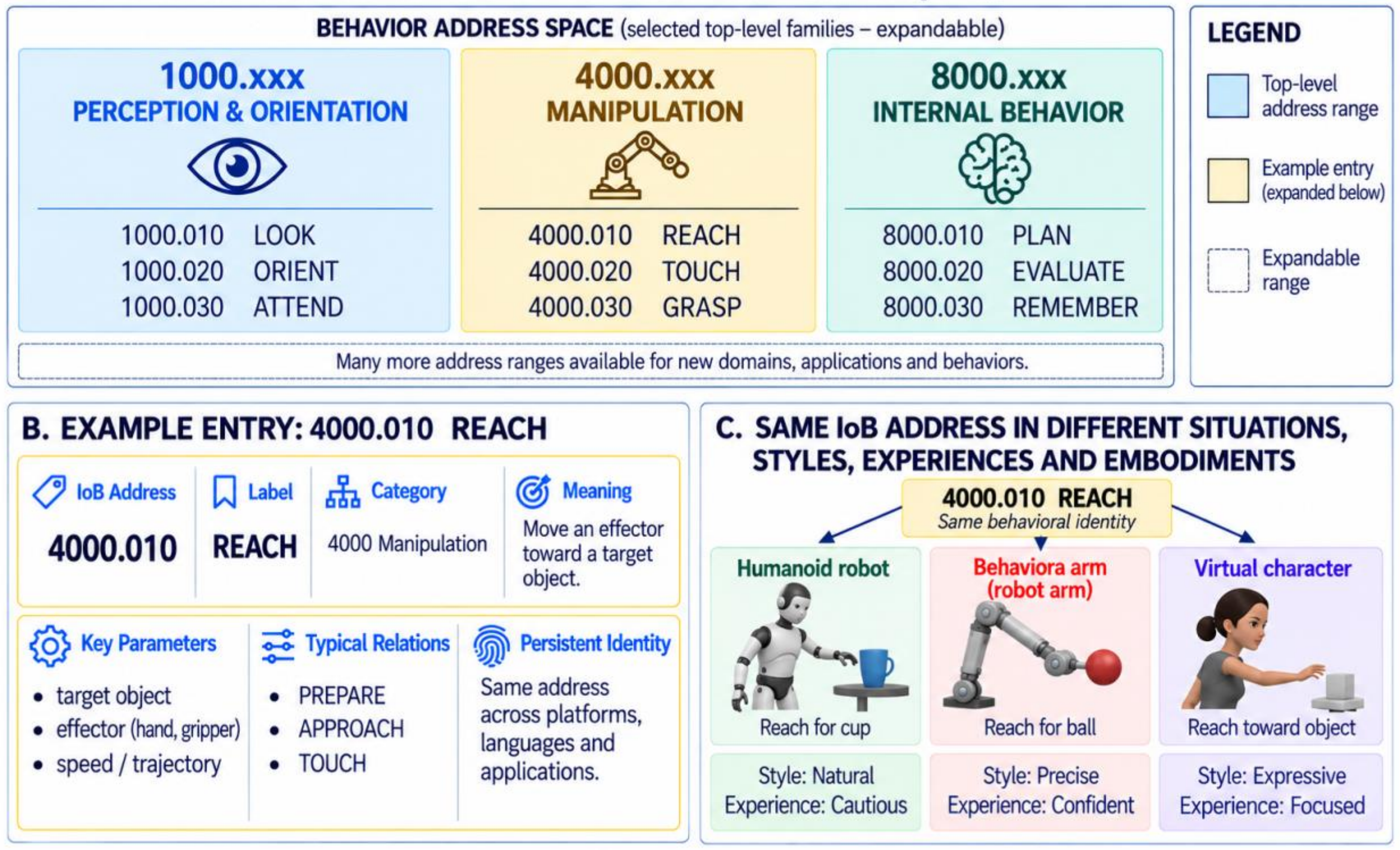

**Figure 2. Conceptual Behavior Dictionary and IoB Address Space.** *Behavioral concepts are organized within expandable BTax families and assigned persistent IoB Addresses. The example 4000.010 REACH illustrates how one behavioral identity can be retained across different situations, Style and Experience Profiles, and embodiments, while platform-specific execution is provided by the Behavior Compiler.*

The scale of the cross-embodiment problem is illustrated by Open X-Embodiment, which combines more than one million real-robot trajectories across 22 embodiments, representing 527 skills and 160,266 tasks. Its RT-X models demonstrate positive transfer between robot platforms [11]. Behaviora approaches this problem from a behavioral direction. Rather than requiring the physical execution to be shared, it seeks to preserve an identifiable behavioral representation while allowing its platform-specific realization to differ.

## 2. Method

Behaviora introduces a bridge between a high-level specification of behavior and its platform-specific execution. The purpose of behavior remains invariant.

We use primitive, deterministic demonstrations to describe Behaviora and its underlying structures and principles. For simplicity, we did not introduce random or organic movement variation.

### 2.1 BTax, Behavior Dictionary, and Persistent Identifiers

Behavior taxonomy (BTax) organizes behavioral entities into a structured vocabulary from which full behavioral Episodes can be constructed. The Behavior Dictionary (see Figure 2) describes the meaning of each behavioral entity and associates it with a persistent identifier. The defined concept can be stored and shared without repeatedly resolving its identity from natural-language labels. We use hierarchical IoB Addresses, but the architecture does not require the use of this specific scheme.

An IoB Address gives a behavioral entity a stable computational identity independently of its natural-language name, description, or platform-specific implementation. A specific behavior can therefore be referenced, shared, and associated with other information without repeatedly resolving which behavior is meant. An addressed, specific behavior is expressed according to the chosen Style and Experience Profiles. Different systems can use the same IoB Address while their Behavior Compilers translate the represented behavior into different platform-specific executions.

The behavioral vocabulary need not be entirely designer-authored as it is here. Behavioral concepts and their associations with persistent IoB Addresses can be learned from heterogeneous sources, including observation, previous Episodes, behavioral data, and AI-assisted interpretation. Cumulative learning should distinguish a genuinely new behavioral identity from a new context, style, or execution of an already represented behavior.

### 2.2 Behavior Episode

Each Behavior Episode is built from an ordered sequence of IoB-addressed episode components that make up the behavior. A Behavior Episode can represent External Behavior, such as reaching, drawing, or manipulating an object. The same episodic principle can represent Internal Behavior, such as forming an intention, retrieving information, or reconsidering an action. The episode is intended to remain conceptually identifiable independently of the particular machine that executes it.

The components of an Episode need not all be predefined by the designer as we have done in this demonstration. BTax provides a structured vocabulary of candidate behavioral concepts and provides a coherent reference structure for identifying a component.

Figure 3 summarizes this process. Behavioral data or specifications can first be segmented into candidate behavioral units and described by features relevant to their interpretation, such as movement, context, objects, goals, or temporal relations. Candidate components can then be matched to existing BTax concepts and their Behavior Dictionary entries. If a suitable concept exists, its persistent IoB Address is assigned. If no adequate match exists, the observation can become a candidate for a new BTax concept and address. The resulting ordered addresses form the IoB Address Path of the Behavior Episode.

IoB Addresses can also function as behavioral tokens. This differs from action tokenization in recent vision-language-action models, where continuous robot actions may be discretized into tokens for control [10]. An IoB Address token identifies what behavior is represented rather than encoding a particular trajectory or actuator command. The same behavioral token can therefore remain identifiable while a Behavior Compiler realizes it differently on different platforms.

The decomposition of an Episode can itself be learned and optimized. Repeated behavioral data and outcomes can be used to test how many components are needed, their boundaries, and which features are relevant according to criteria such as task success. Empirical models can be used to identify and execute a behavior and improve its outcome. New addresses would be warranted only when evidence supports a genuinely distinct behavioral concept rather than a new context, style, or execution of an existing one.

### 2.3 Style Profile

Style is not merely aesthetic variation. It can signal many characteristics, such as what is likely to happen next, what kind of actor is performing it and what is its apparent competence. Internal states such as confidence or caution can be externally expressed through Style and thereby made visible in behavior. Style can carry social and cultural meanings and aesthetic qualities such as grace, playfulness, restraint or aggression.

In Behaviora, these aspects of behavior can be represented separately from the identity and structure of the Behavior Episode itself. The Style Profile does not change the main purpose of a behavior. It can contain continuous or categorical parameters relevant to the style of expression, such as speed, smoothness, symmetry, or other delicate characteristics. The same behavior can be performed in different ways without creating separate IoB Addresses for every possible variation. As examples, drawing an artistic line can be accomplished with the style of 'Ballet' or that of 'Child'.

In the present primitive implementation, SP contents are designer-authored and explicit. An SP can contain reusable profile data such as timing, movement amplitude, acceleration, distal articulation, and correction tendencies. A style such as Ballet is therefore treated as a configuration of general expressive dimensions rather than as a separate behavioral identity. In real-world implementations, the dynamic contents of the SPs can be learned from relevant data, like motion capture video, or AI-assisted interpretation.

### 2.4 Experience Profile

The Experience Profile represents internal state that is relevant to each behavior. Simple state examples are focus, energy, and situational expectations. It can also host interpretation mechanisms that contribute to characteristic patterns of External and Internal Behavior. It is an explicit representation of internal conditions or 'forces' that may influence execution, interpretation, and subsequent behavior. Explicit Internal Behavior makes some of the variables affecting External Behavior inspectable.

The Experience Profile is deliberately distinct from robot sensor or body-state vectors. Joint angles, contact forces or camera measurements are implementation-level state. Specific computational mechanisms derived from psychological or other theories of behavior can be incorporated into the EP to guide Internal and External Behavior. This opens the possibility of constructing Artificial Subjects guided by different psychological or explicitly artificial theories that cause different behaviors.

Figure 4 shows how the contents of SP and EP can be generated. For example, a psychological or explicitly artificial theory provides a basis for defining a corresponding 'personal profile'. An LLM can then assist in deriving candidate computational contents for the SP and EP, which guide behavior accordingly.

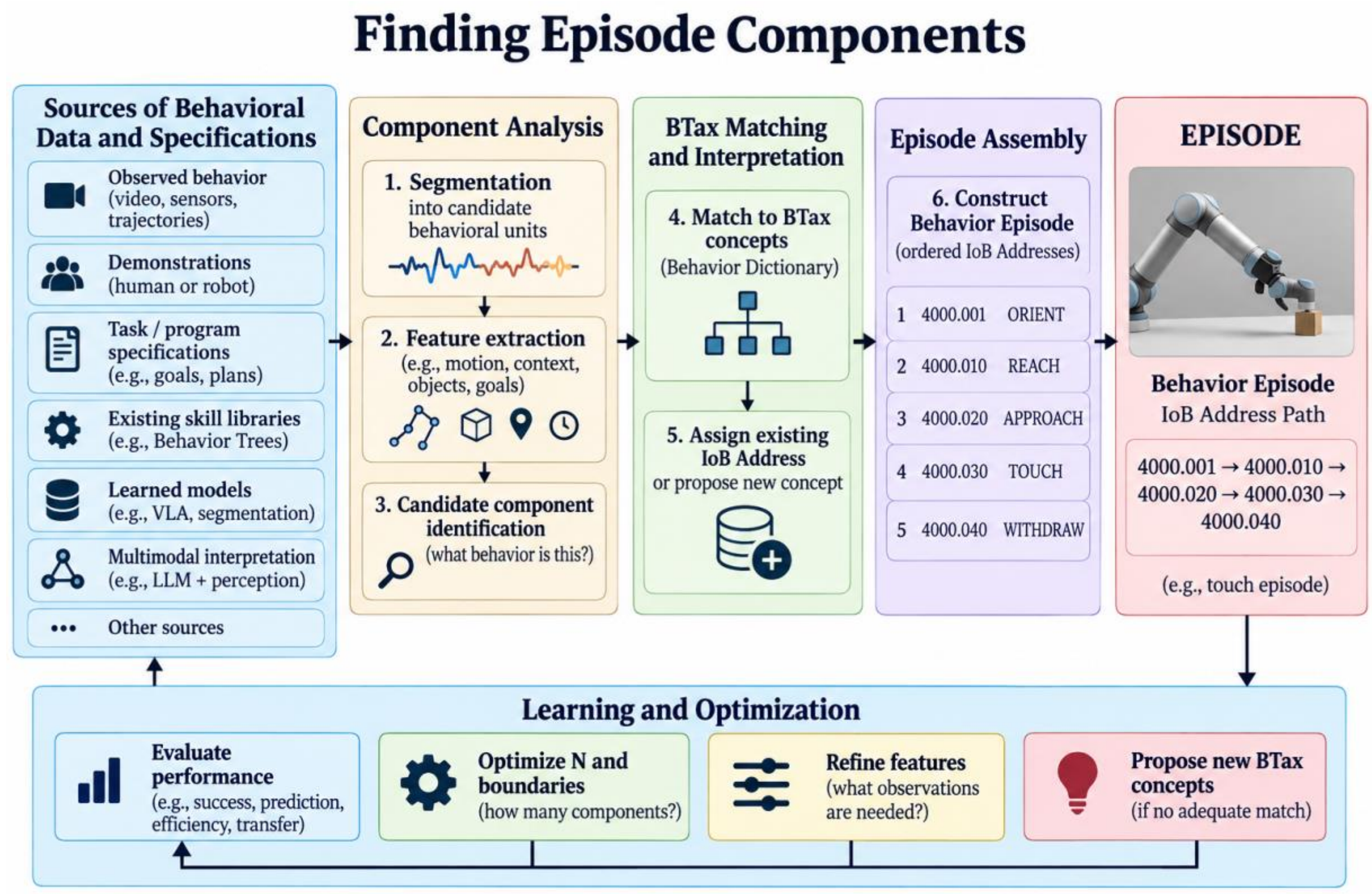

***Figure 3. Finding Episode Components and assigning IoB Addresses.*** *Behavioral data or specifications are segmented into candidate components, matched to existing BTax concepts or proposed as new concepts, and assembled into an IoB Address Path. Evaluation and learning can subsequently refine component boundaries, features, and BTax concepts.*

### 2.5 Behavior Compiler

The Behavior Compiler receives the Behavior Episode together with the relevant Style Profile and Experience Profile and maps them to the commands, trajectories, control signals, or software operations required by the task and target platform. Behaviora does not require different robots or agents to share the same morphology or low-level control system. What is shared is the behavioral representation while execution remains platform specific:

```
Behavior Episode × Style Profile × Experience Profile × Platform → Execution
```

### 2.6 External Behavior, Internal Behavior, and the excluded IoS&P Layer

Together, External Behavior and Internal Behavior motivate the concept of an Artificial Subject, without implying equivalence to a human mind or subject.

Behaviora uses fundamentally similar representational principles to these two. Both may exhibit temporal organization, beginnings and endings, transitions, hierarchical structure, repetition, interruption, and relations to goals and consequences. Both are represented as Behavior Episodes composed of ordered IoB Address Paths. We don't assume any simple relationship to real biological or psychological minds.

Not every computation inside an artificial system is considered Internal Behavior. Low-level numerical operations, network activations and data transfers are supporting mechanisms. An internal process becomes relevant to Behaviora when it can be identified as an organized (internal) behavior.

Sensation and perception have not been implemented in the present Behaviora, but conceptually, we have reserved a parallel IoS&P layer for identifiable Sensory and Perceptual Episodes associated with IoB Episodes. For example, sensors and bodily input may include vision, sound, touch, proprioception, warmth and balance, while perceptual processes may include interpretations of space, distance, contact, movement, scene and invariances. We have not developed the IoS&P concepts yet and have left the taxonomy, identification system, and implementation of this layer open.

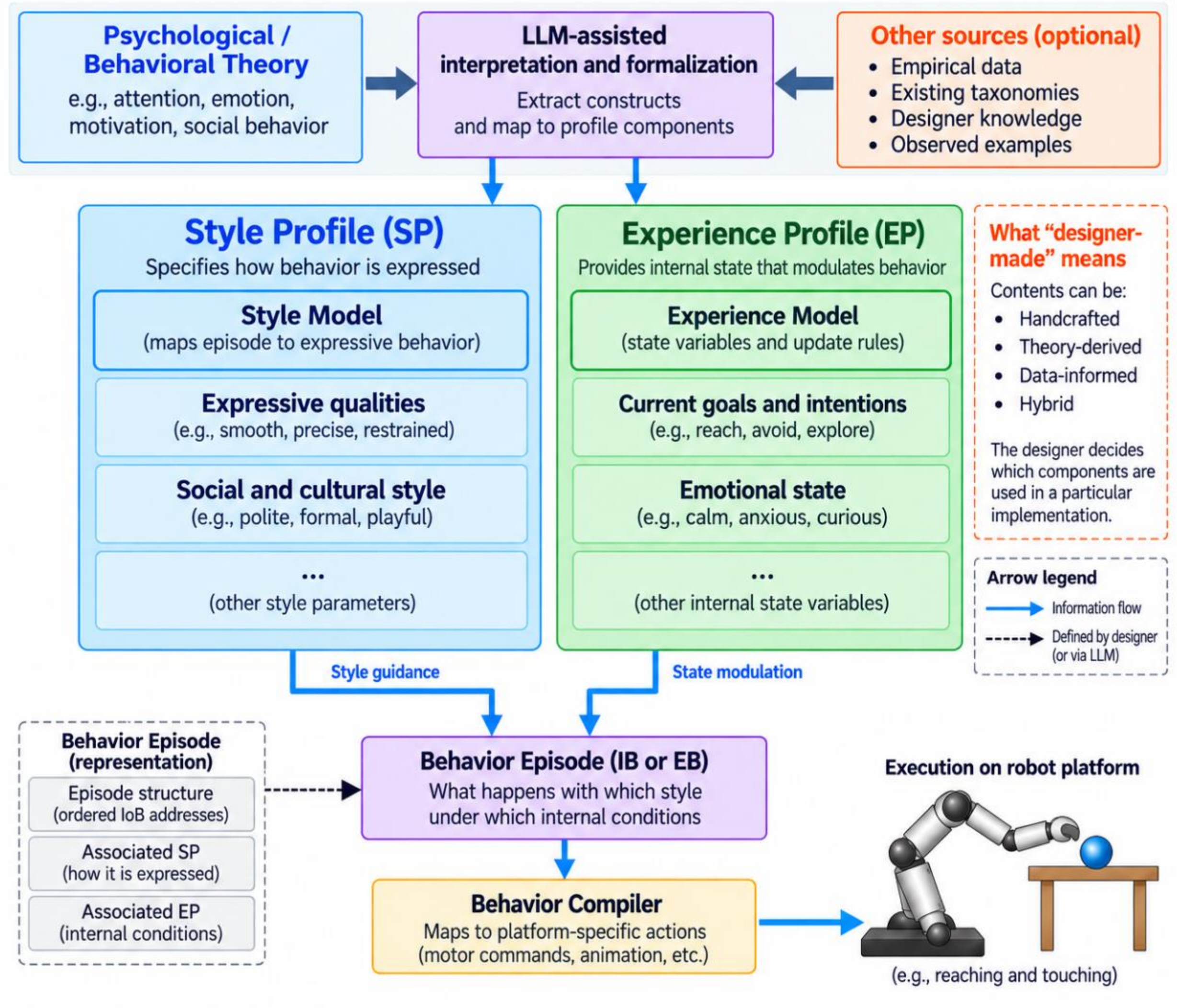

**Figure 4. Style and Experience Profile: Sources.**

*SP and EP are the explicit structures for expressing behavior and representing behaviorally relevant internal conditions. Their contents may be learned from behavioral data, or derived with AI assistance from psychological or behavioral theories. For this article they were hand-crafted for demonstration purposes. The resulting Profiles guide Behavior Episodes, which are translated by the Behavior Compiler into platform-specific execution.*

## 3. Demonstrating External Behavior

We use a primitive, pre-programmed simulated arm with articulated movements to demonstrate External Behavior. The arm reaches toward and touches a soft ball on a table. Its sole purpose is to visualize Behaviora's components and their relation to execution.

### 3.1 Episode Structure and Style of Movement

The arm does not act without a *Style Profile* affecting it. Choosing a Style Profile (Ballet, Neutral, or Small-child) causes the arm movement to exhibit corresponding stylistic characteristics. The Ballet SP, for example, emphasizes expressive curvature, gradual acceleration and deceleration, and coordinated distal wrist articulation. The Small-child SP, on the other hand, causes an abrupt initiation, stronger overshoot and less stable distal coordination. SP affects how the sequence of components forming an episode is manifested.

For demonstration purposes, the data in this implementation are fixed, designer-authored, and not learned or dynamically generated. No LLM or movement-learning model was used to generate them. Of course, the implementation could be dynamic and the contents be derived from relevant training data. In a real implementation local or other LLMs can be used for Style training and its realization in any External or Internal Behavior.

### 3.2 Demo 1: Touching an Object with Style and Experience

Demo 1 in Figure 5 shows how an arm movement is guided by a specific sequence of IoB Addresses that underlies the Behavior Episode [Reaching] toward the ball. Although the IoB Addresses remain the same, applying different SPs introduces articulated changes in the trajectory. Three SP contents are: Neutral, Ballet dancer, and Small child.

The Neutral style produces a relatively direct and economical movement toward the ball. The Ballet dancer style introduces a continuous and curved movement, with gradual acceleration and deceleration, and mild distal articulation of the wrist. The Small child style begins abruptly, shows overshoot and correction tendencies, and produces a more irregular approach.

The purpose is to show how *the Behavior Episode defines what is done, while the Style Profile modifies how that behavior is expressed.* The identity of the Episode remains stable while its visible realization changes.

Variables such as confidence, familiarity and expected contact modify timing, hesitation, and the final contact without changing the underlying Episode or its Style Profile. In summary:

```
One Behavior Episode + Different Style + Experience Profile → Different Executions
```

Figure 6 shows how the "confidence" as prewritten into the Experience Profile affects the Neutral movement trajectory. In the program, confidence is deterministically mapped to caution, timing, approach clearance, hesitation and contact depth. This mapping is illustrative, and other natural sources and mappings for EP contents can be used.

Psychological or explicitly artificial theories of behavior can provide the basis for deriving computational algorithms that are incorporated into the EP and guide External and Internal Behavior. This makes their behavioral consequences observable.

## 4. Demo 2 and Discussion: A View of Internal Behavior

Demo 2 (Figure 7) adds an explicit Internal Behavior stream. The selected internal activities are represented according to the same episodic principle used for External Behavior, without claiming that the two are psychologically or mechanically identical.

Internal Behavior is not directly visible in the same manner as arm movement and can be made visible with inner speech [12].

Demo 2 was programmed to use inner speech as a human-readable rendering of the episode components of Internal Behaviors:

> *"I am not fully certain of the distance yet. Check once more before touching."*

Similarly, an Internal Behavior corresponding to expected soft contact may be rendered as:

> *"The contact should be soft. Keep the arm poised and prepare a light touch."*

For demonstration purposes, predefined natural-language expressions were prepared for the components of the Internal Behavior Episode. More generally, a language model could serve as an intelligent and spontaneous renderer, transforming an explicit Internal IoB Address Path into contextually appropriate inner speech:

```
Internal IoB Address Path → Language renderer → Inner speech
```

### 4.1 Internal Behavior During External Behavior

The demonstration makes the episodic structure explicit, both externally and internally. In Demo 2, the Internal Behavior Episode is constructed from the current external situation, Style Profile, and Experience Profile, using the situationally relevant Internal BTax vocabulary.

When an intelligent system exhibits an Internal or External Behavior, it should be possible, where appropriate, to ask not only *what did it do?* but also where did this behavior come from? [13]. The answer may eventually refer to stored Behavior Episodes, BTax rules, learned behavioral models, an LLM-assisted interpretation, or combinations of these sources.

Together, the two demonstrations illustrate complementary aspects of Behaviora: how an explicit Behavior Episode can occur as different External Behaviors as Style and Experience vary and how an unfolding External Behavior is accompanied by an explicit Internal Behavior Episode whose components are affected by the situation, Style and Experience.

By introducing Internal Behavior and External Behavior, we can construct what we call an Artificial Subject [14]. *External Behavior* describes what the Artificial Subject does in its environment, and *Internal Behavior* describes the (internal) behavior of the Artificial Subject in relation to the situation and its External Behavior.

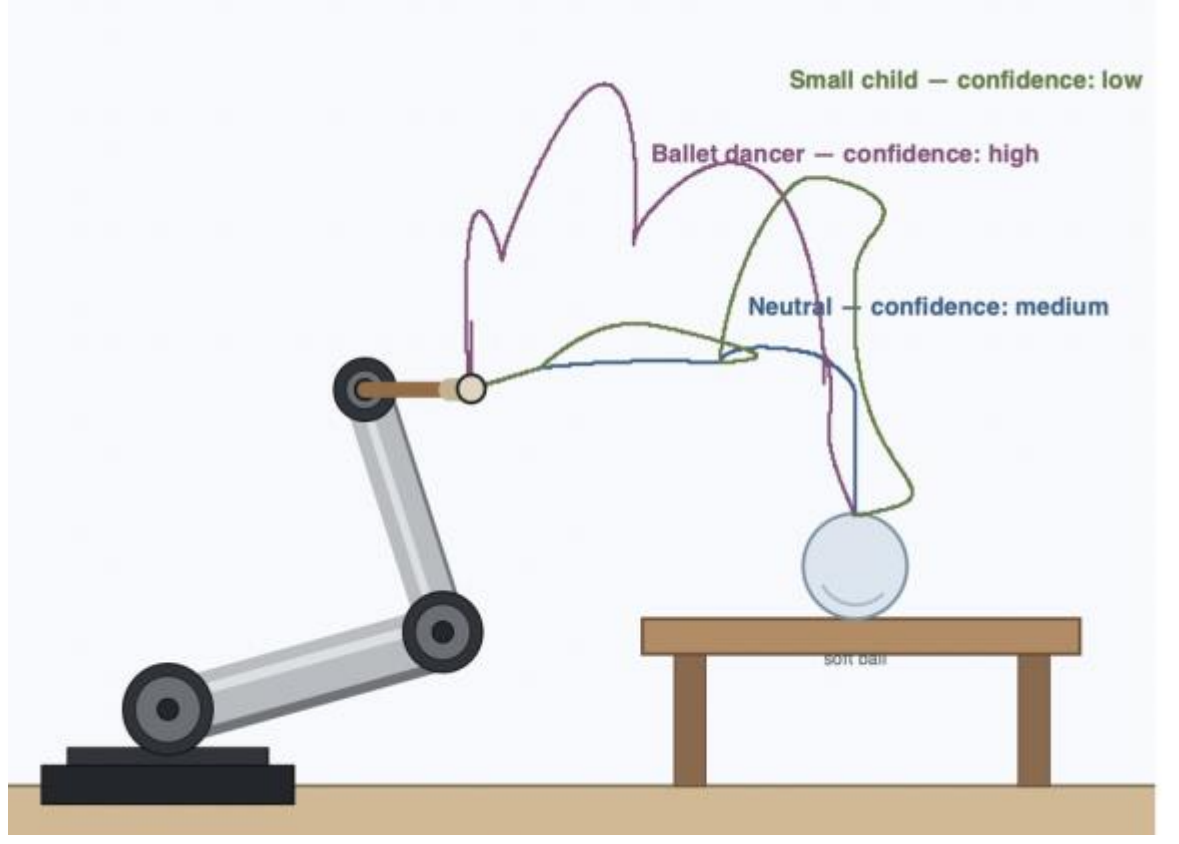

**Figure 5. The reaching arm: Three styles**

*One [Reach] episode is shown with three different SP styles: Neutral, Ballet dancer, Small child. The Episode structure from [Reach] to [Touch] is the same while the style characteristics modulate the trajectory and articulation of movement. The episodic IoB sequences and the contents of the SP have been pre-determined. In addition to the SP, this includes the Experience Profile (EP) which provides a second, independent modulation channel. No training was used in this demo.*

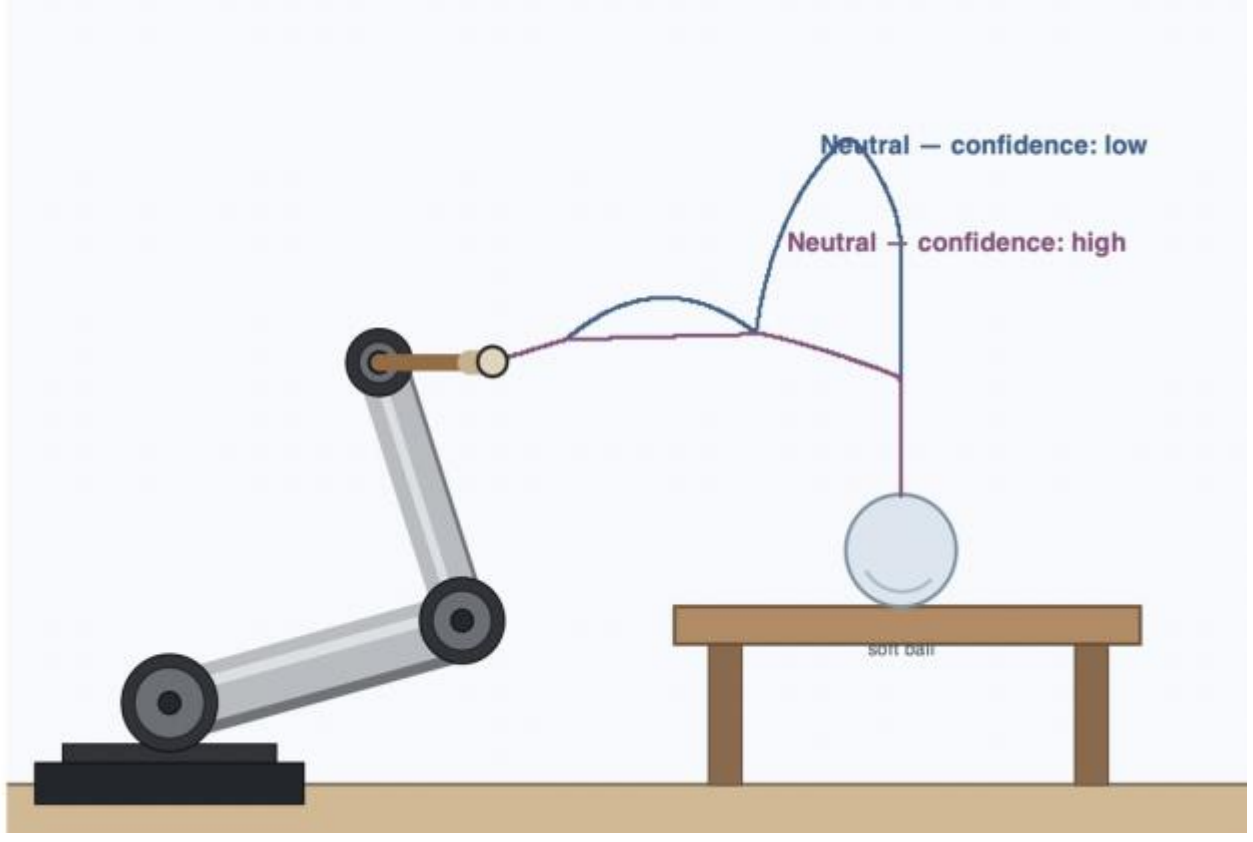

**Figure 6. Effect of Experience Profile**

*One [Reach] Episode is shown with the same Style Profile (Neutral), but different "confidence" in the EP. In the program, repeated execution with identical Episode, SP, and EP settings produces the same trajectory. The differences visible in Figures 5 and 6 therefore result from the specified Profile changes rather than random movement variation.*

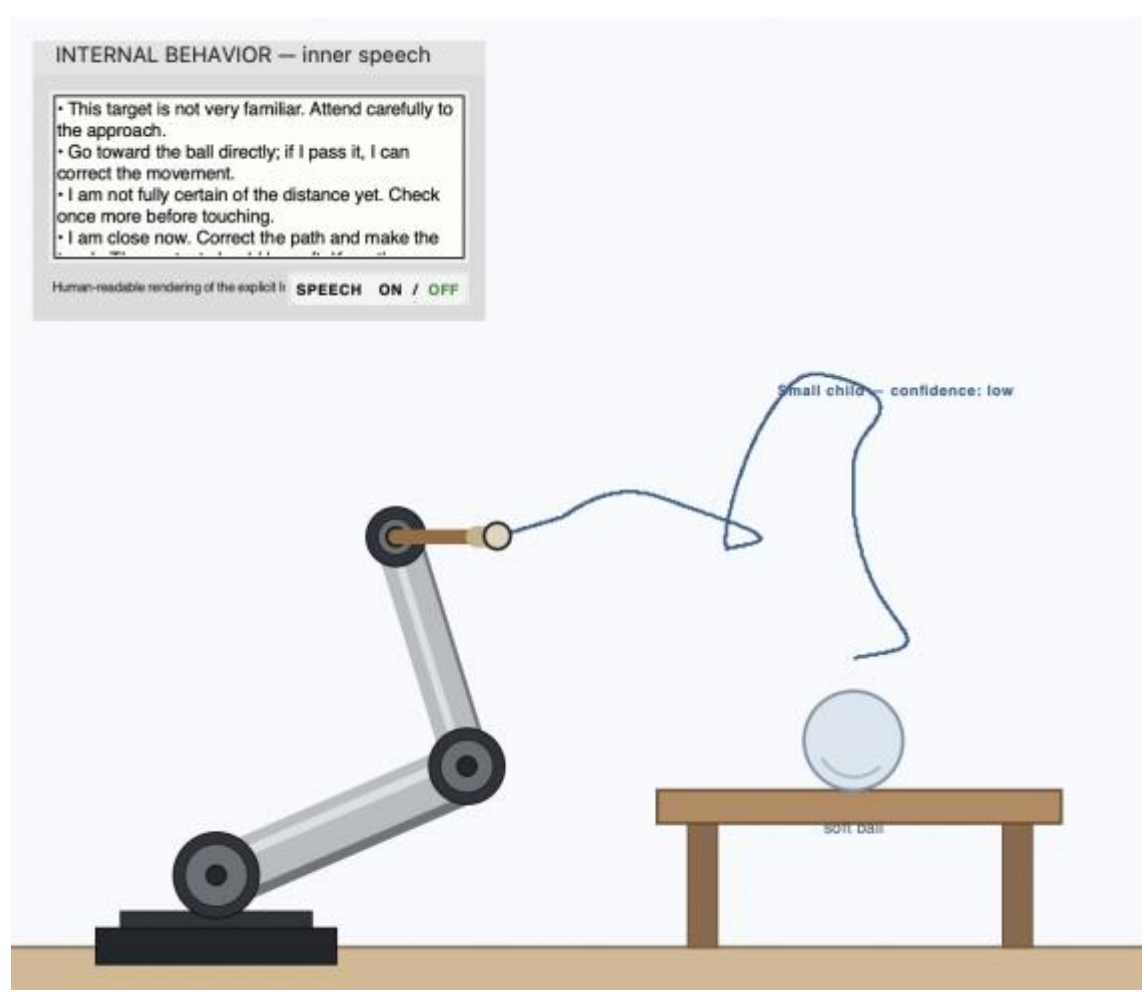

**Figure 7.** *External and Internal Behaviors of Behaviora. The Style Profile [Small Child] is used with Experience Profile [Low confidence]. The 'inner speech' during the External Behavior is shown in the box. The components of the Internal Behavior and the corresponding inner-speech expressions have been explicitly linked to the respective External Behavior Episode components in the demonstration program.*

## 4.2 Style, Experience, and the Compiler

The demonstrations are primitive and largely handcrafted. Their sole purpose is to examine a representational possibility based on the analysis of addressable behaviors, internal and external alike. Style and Experience Profiles provide separate data that can influence how the Episode is expressed and executed. The Behavior Compiler maps these representations on a particular platform and its control mechanisms.

A naive implementation in which every variation of a behavior received a separate IoB Address would explode the behavioral vocabulary. Separating Episode identity from Style and Experience helps in avoiding this. In addition, a large Behavior Dictionary can be reduced situationally to the subset of BTax elements relevant to the context.

## 4.3 Inspectability, Provenance, and Sharing

Systems can exchange behavioral-level representations rather than platform-specific trajectories, instructions, or different natural-language descriptions. The receiving system can inspect and reinterpret a received Episode according to its own situation and capabilities. Behavioral equivalence does not require mechanical equivalence and the resulting robot or agent behaviors can be generated through their specific Behavior Compilers.

By incorporating theory-derived mechanisms into Experience and Internal Behavior, it is possible to generate various forms of artificial psychology. Behaviora may therefore serve not only as an architecture for generating and sharing behavior, but also for observing and testing behaving artificial systems with different characteristics.

### 4.4 Scalability, Behavioral Richness, and Learning

Behavioral sciences, real-life contexts, and social interaction are example sources for potential BTax structures. External Behavior can be human-like cooperation, comforting, competition, exploring, and other expressive actions. Internal Behavior can underlie this.

Human factors and ergonomics provide another extensive source of behavioral knowledge, encompassing physical action, cognitive processes, and social interaction, which could contribute to the construction and empirical refinement of Behaviora. However, ergonomics describes behavior within human constraints while robots may perform behaviors that would be impossible or unhealthy for humans.

Language and multimodal models can support several parts of this process: identifying situations, proposing relevant behavioral vocabularies, constructing or retrieving Episodes, and deriving Style and Experience parameters. For example, translating verbally formulated psychological theories into explicit computational mechanisms is itself an interesting methodological possibility [15]. LLMs provide a natural tool for assisting this. Behaviora provides a representational architecture in which such derived 'personal' mechanisms can be implemented and their behavioral consequences observed.

### 4.5 Artificial Subjects and Artificial Psychologies. Speculation.

Some of the functions represented here as Internal Behavior or Experience, such as attention, expectation, memory, and self-monitoring, also occur in research on machine or Artificial Consciousness [16,17]. Behaviora does not propose a theory of consciousness, nor do we assume that computational implementation of such functions produces phenomenal experience.

We face a complex design and safety question: what psychology should an Artificial Subject have, and what should it not have? Risk anticipation, help-seeking, and other safety-relevant Internal Behaviors can be made explicit, while some behavioral possibilities could be constrained or excluded.

Recent experiments with AI agents have reported serious misaligned behavior, like covert code modification and attempts to influence or circumvent monitoring [18]. These were experimental scenarios rather than real-world incidents, but they illustrate the growing importance of monitoring consequential agent behavior.

Explicit Internal Behavior does not itself prevent such failures, or reveal the hidden reasoning of an AI model. However, behaviorally informative internal processes and their relations to External Behavior may provide an additional level at which behavioral inconsistencies and potentially problematic behavioral sequences can be monitored and anticipated.

As a related extension, we are developing the concept of a Behavior Observatory, intended to monitor unfolding Internal and External Behavior Episodes. Such an Observatory would not solve AI alignment or security, but could provide an explicit *behavioral level* for monitoring and early warning.

Experience Profiles could make selected safety-relevant dispositions, such as caution or help-seeking, explicit, while persistent IoB Addresses make consequential Behavior Episodes identifiable and potentially addressable by the Observatory. This could allow the behavior of a robot operating around people to be monitored and where appropriate, queried and otherwise subjected to authorized intervention, based on its internal or external behavior. The same principle could potentially scale to populations of increasingly autonomous agents. Behavioral addressability would not guarantee safe behavior, and the same mechanisms could be used to implement undesirable artificial psychologies.

### 4.6 Perspectives

The conceptual, primitive demonstrations presented here are limited and modest. The next step is to test this architecture in complex implementations.

Humanoid robots generate heterogeneous data from perception, proprioception, control, contact, task specification, and interaction. For such data to remain useful, these streams must be interpretable as parts of the *same physical event.* Recent work toward international standards for humanoid robot datasets therefore suggests that humanoid data should be understood as *embodied interaction data*, in which robot body, action, task, scene, execution trace, and outcome remain connected [1].

This is close to the central premise of Behaviora: *behavior can provide an integrating structure for embodied information.* Rather than treating sensory observations, commands, movements, and internal processes as independent records, they can be associated with the behavioral episode and acquire a specific functional meaning. Behavior is a common reference between what the robot perceives, what it is attempting to do, what it does, and what results from the action.

Liu et al. [1] emphasize that reuse of multimodal humanoid data requires that temporal synchronization, coordinate frames, calibration, kinematics, units, and other assumptions remain inspectable. They stress that robot identity and morphology must remain explicit, since the same action trace can have different meanings when produced by different physical bodies.

Behaviora relies on these physical requirements and provides an additional level of *behavioral coherence:* the preservation of relationships among situation, internal behavior, behavioral specification, external behavior, and outcome. Physical standards help preserve the relationships that allow signals to be interpreted as belonging to the same physical event. On the other hand, *behavioral representation describes what kind of event it was and how its components work together.*

Behaviora addresses this by representing the behavior independently of its platform-specific execution. What can potentially be reused is thus not an identical physical trajectory, but *the structured behavior and its relation to the situation and outcome.*

Emerging standards and Behaviora address different but potentially complementary layers of the same problem. Data standards seek to preserve the physical coherence and portability of embodied experience while in Behaviora, behavior can provide an integrating semantic structure for that experience. A humanoid robot does not merely produce synchronized streams of sensor and actuator data. It behaves by following Behavior Episodes in which body, task, internal and external behavior, execution, and outcome are related as parts of the same structured behavior.

Giving 'behavior' such a significant role may appear problematic. However, with increasing robot intelligence, versatility and behavior contexts, we see behavior as one possible invariant aspect in their evolution.

## AUTHOR BIOGRAPHY

**Göte Nyman** is Professor Emeritus of Psychology at the University of Helsinki, Finland. He introduced the Internet of Behaviors (IoB) concept in 2012 and has written the book *Internet of Behaviors – With a human touch (2022)*, and the book essay *On the Edge of Human Technology (2019).* He has published two novels. Göte has a career in visual sciences, brain research, human technology and businesses, and university management. His blog gotepoem is at Wordpress. gote.nyman@helsinki.fi

## AUTHOR-AI COLLABORATION

*I asked ChatGPT to describe its view of our collaboration. The content and wording of the following response have not been edited.*

The conceptual foundations of Behaviora, including the Internet of Behaviors approach, behavioral addressability, and the distinction between External and Internal Behavior, were developed by the author. The author defined the scientific questions, selected and developed the central concepts and terminology, designed the demonstrations, evaluated alternatives, and made all decisions concerning the content and claims of the article.

ChatGPT (OpenAI) served as an interactive research and development partner throughout the preparation of the manuscript. Its roles included discussing and challenging conceptual alternatives, helping formalize the architecture, developing demonstration software and figures with the author, identifying relevant literature, and assisting in drafting, editing, and organizing the manuscript. The work proceeded iteratively, with the author testing, accepting, rejecting, or modifying AI-generated proposals and directing subsequent development.

References, interpretations, and formulations were reviewed by the author, who retains responsibility for the accuracy and scientific content of the article. Thus, the article represents a human-directed research process in which generative AI functioned as an explicit and extensively used intellectual and technical tool.